\documentclass{article}
\usepackage{ijcai24}

\usepackage{times}
\usepackage{soul}
\usepackage{url}
\usepackage[hidelinks]{hyperref}
\usepackage[utf8]{inputenc}
\usepackage[small]{caption}
\usepackage{graphicx}
\usepackage{amsmath}
\usepackage{amsthm}
\usepackage{booktabs}
\usepackage{algorithm}
\usepackage{algorithmic}
\usepackage[switch]{lineno}

\usepackage{color}
\usepackage{bbding}
\usepackage{amsmath}
\usepackage{amsfonts}
\usepackage{multirow}
\usepackage{booktabs}
\usepackage{tabularx}
\usepackage{subfigure}
\usepackage[para,online,flushleft]{threeparttable}

\title{Query-Based Knowledge Sharing for Open-Vocabulary Multi-Label Classification}

\author{
Xuelin Zhu$^1$\and
Jian Liu$^2$\and
Dongqi Tang$^{2}$\and
Jiawei Ge$^1$\and
Weijia Liu$^1$\and
Bo Liu$^1$\and
Jiuxin Cao$^1$\\
\affiliations
$^1$Southeast University\\
$^2$Ant Group\\
\emails
\{zhuxuelin, jiawei\_ge, weijia-liu, bliu, jx.cao\}@seu.edu.cn,
\{rex.lj, dongqi.tdq\}@antgroup.com
}

\begin{document}

\maketitle

\begin{abstract}
Identifying labels that did not appear during training, known as multi-label zero-shot learning, is a non-trivial task in computer vision. To this end, recent studies have attempted to explore the multi-modal knowledge of vision-language pre-training (VLP) models by knowledge distillation, allowing to recognize unseen labels in an open-vocabulary manner. However, experimental evidence shows that knowledge distillation is suboptimal and provides limited performance gain in unseen label prediction. In this paper, a novel query-based knowledge sharing paradigm is proposed to explore the multi-modal knowledge from the pretrained VLP model for open-vocabulary multi-label classification. Specifically, a set of learnable label-agnostic query tokens is trained to extract critical vision knowledge from the input image, and further shared across all labels, allowing them to select tokens of interest as visual clues for recognition. Besides, we propose an effective prompt pool for robust label embedding, and reformulate the standard ranking learning into a form of classification to allow the magnitude of feature vectors for matching, which both significantly benefit label recognition. Experimental results show that our framework significantly outperforms state-of-the-art methods on zero-shot task by 5.9\% and 4.5\% in mAP on the NUS-WIDE and Open Images, respectively.
\end{abstract}

\section{Introduction}
Multi-label image classification is a fundamental task in the field of computer vision that focuses on identifying multiple objects and concepts, also known as labels, within images. In a typical scenario, the candidate set of labels on both the training and testing phases are the same. However, real-world applications often present challenges as new and unanticipated labels may emerge, particularly given the limited size of the label set in the training data. As such, the ability to accurately recognize previously unseen labels at test time independent of their annotated training instances is an important issue currently under consideration.


In order to recognize unseen labels in images, most of the existing methods \cite{huynh2020shared,ben2021semantic,narayan2021discriminative} resort to zero-shot learning (ZSL), which commonly employs pretrained language models like Glove \cite{pennington2014glove} to transfer pretrained knowledge from seen labels to unseen labels, thus making the recognition of unseen labels feasible. However, these multi-label ZSL methods only explore the knowledge transfer in text modality, while ignore informative visual modality as well as their cross-modal semantic knowledge, providing poor performance that are far away from satisfying the requirement of practical applications.


\begin{figure}[t]
    \centering
    \subfigure[MKT]{
        \includegraphics[width=0.58\linewidth]{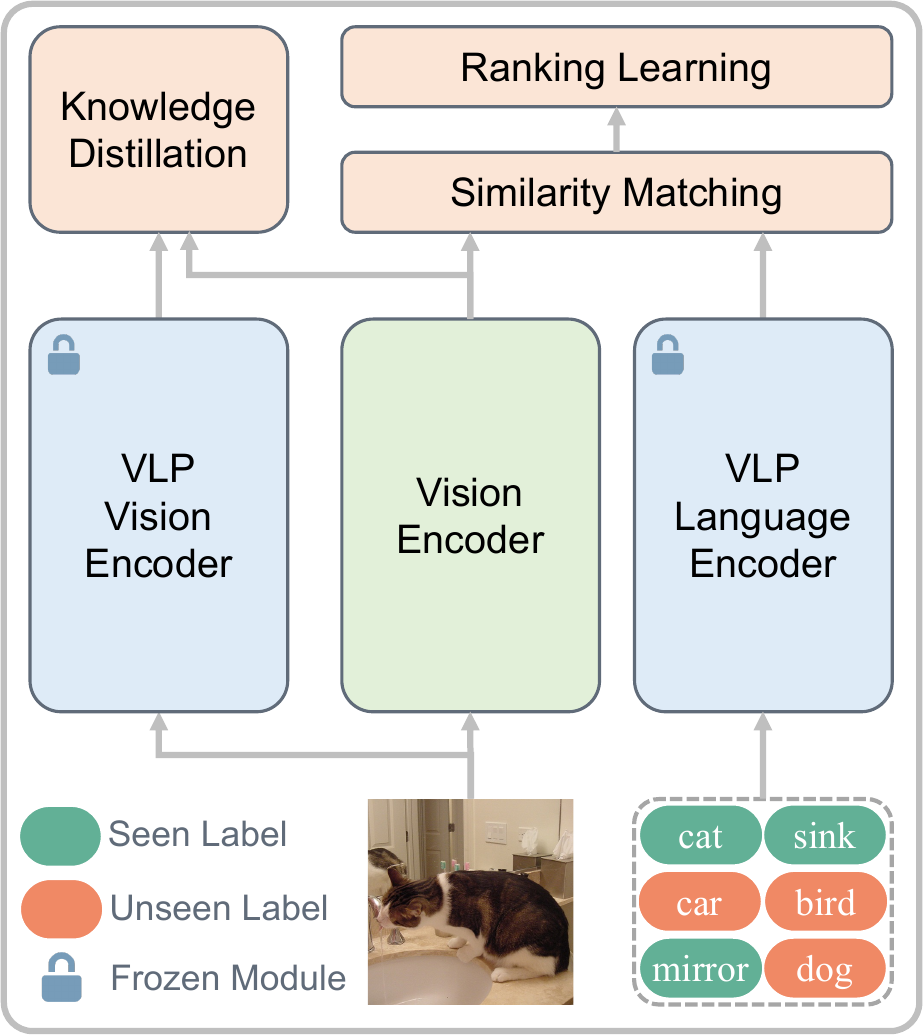}
        \label{fig:mkt}
    }
    \hskip -1.2ex
    \subfigure[QKS]{
	  \includegraphics[width=0.38\linewidth]{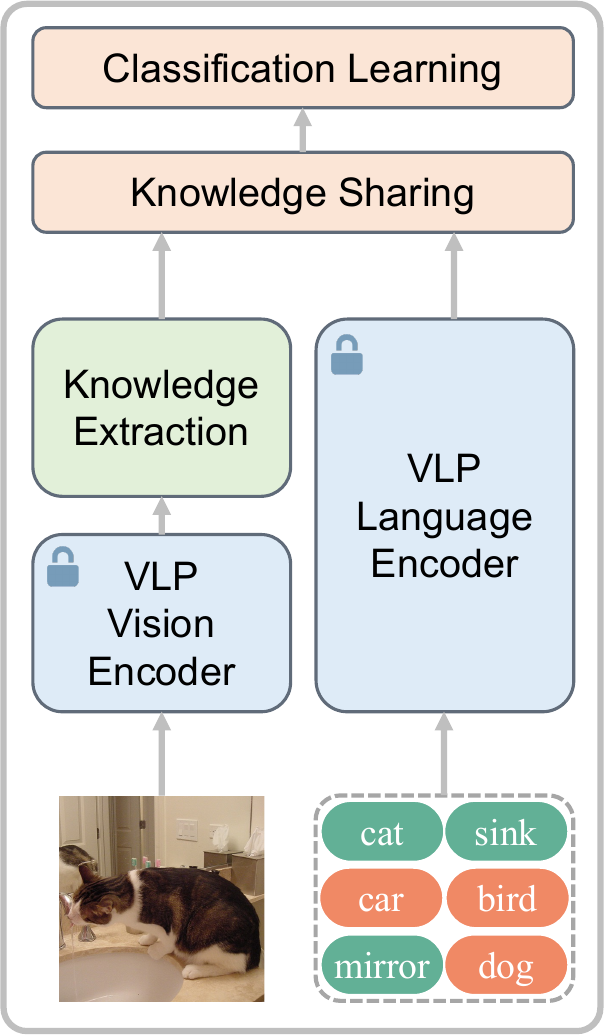}
        \label{fig:qks}
    }
    \caption{A brief comparison on paradigms of exploring pretrained vision-language models for the open-vocabulary multi-label classification task. (a) MKT relies on knowledge distillation to preserve the image-text matching ability of the VLP model and performs ranking learning for label recognition. (b) Our QKS takes the VLP model as part of the framework and designs a vision knowledge extraction module to explore crucial and informative vision features for matching with label embeddings by classification learning.}
    \label{fig:intro}
\end{figure}



Benefiting from the pretraining with millions of image-text pairs, VLP models have acquired multi-modal knowledge of general concepts. Exploring such knowledge for label recognition, known as open-vocabulary multi-label classification, has become increasingly popular. MKT \cite{he2023open}, a best existing method, uses knowledge distillation to preserve the image-text matching ability of VLP models, thus enabling multi-modal knowledge transfer for unseen label recognition, as shown in Figure \ref{fig:mkt}. Despite its success, we argue that knowledge distillation is suboptimal in terms of exploring the multi-modal knowledge of VLP models. According to the Table \ref{tab:nuswide} and Table \ref{tab:openimage}, MKT exhibits only a slight performance advantage in mAP, or even downsides in F1 score, over the plain CLIP \cite{radford2021learning}. A major reason is the fact that CLIP struggles with excessive polysemy \cite{abdelfattah2023cdul}, which greatly weakens the effect of knowledge distillation in the multi-label task. Besides, MKT requires two-stage training, which is time-consuming and labor-intensive. It also suffers from redundant visual information in global features since labels are usually only relevant to particular regions of images. Overall, an effective paradigm to fully explore the multi-modal knowledge of VLP models for label recognition still remains to be established.

In light of the above analysis, we propose a novel \underline{Q}uery-based \underline{K}nowledge \underline{S}haring (QKS) framework for open-vocabulary multi-label classification. As illustrated in Figure \ref{fig:qks}, our QKS incorporates the VLP model as the foundation of the whole framework, followed by a knowledge extraction module and a knowledge sharing module. Instead of using knowledge distillation, we freeze the whole VLP model to preserve its pre-training multi-modal knowledge and employ it to encode the spatial features of the input image and semantic embeddings of the prompted labels. Then, in the knowledge extraction module, a fixed number of learnable label-agnostic query tokens are trained to aggregate crucial and informative knowledge from the spatial features, filtering out redundant visual information. These tokens full of visual knowledge are subsequently input into the knowledge sharing module and shared across all labels, allowing them to select tokens of interest as key visual clues for label recognition. Besides, we propose two effective techniques, namely prompt pool for enhancing the robustness of label embeddings and ranking as classification to allow the magnitude of feature vectors for image-label matching, which both substantially boost the performance of unseen label recognition. To the end, our main contributions are summarized as follows:

\begin{itemize}
    \item We design a novel knowledge extraction module, which is capable of exploring multi-modal knowledge from VLP models and extracting crucial visual clues for matching with label embeddings.
    \item We propose a simple yet effective prompt technique for label embedding, which provides rich and diverse contexts for each label and yields robust label embeddings for matching with visual features.
    \item We modify ranking learning into a form of classification to enable the magnitude of feature vectors for label prediction, which significantly improves model's performance in precision and recall as well as F1 score.
    \item We propose an effective query-based knowledge sharing paradigm to explore multi-modal knowledge from the pretrained VLP model for open-vocabulary multi-label recognition, which outperforms state-of-the-art methods by 5.9\% and 4.5\% in mAP on the NUS-WIDE and Open Images datasets, respectively.
\end{itemize}

\section{Related Work}

Multi-label zero-shot learning is a cross task of multi-label image classification and zero-shot learning in computer vision, thereby encountering the challenges of both fields. To identify unseen labels, an important idea is to establish their relationship with seen labels. HierSE \cite{li2015zero} uses WordNet for hierarchical embedding representation of label semantics, while Lee \textit{et al.} \cite{lee2018multi} builds label knowledge graphs based on WordNet to model the inter-dependencies between seen and unseen labels. Fu \textit{et al.} \cite{fu2015transductive} exploits the correlations and the unique compositionality property of semantic word vectors, enabling the regression model learned from seen labels to generalise well to unseen labels. Fast0Tag \cite{zhang2016fast} and ZS-SDL \cite{ben2021semantic} aims to estimate the principal directions of images for the purpose of ranking relevant labels ahead of irrelevant labels. CLF \cite{gupta2021generative} uses generative adversarial networks to synthesize multi-label features by label embeddings. However, these methods commonly utilize global image features and provide limited performance in unseen label recognition task. 

Recently, region-based methods for label recognition have received much attention. MIVSE \cite{ren2017multiple} characterizes the region-to-label correspondence by discovering and mapping semantically meaningful image regions to the corresponding labels. Deep0Tag \cite{rahman2019deep0tag} integrates automatic patch discovery, feature aggregation and semantic domain projection within a single uniﬁed framework. LESA \cite{huynh2020shared} proposes a shared attention scheme, which takes a shared approach towards attending to region features with a common set of attention maps for all the labels. BiAM \cite{narayan2021discriminative} utilizes a bi-level attention module to contextualize and enrich the region features, and generate discriminative representations for unseen label prediction. However, these methods exploit only single-modal knowledge using pre-trained language models and their performance is still unsatisfactory.


\begin{figure*}[t]
	\centering
	\includegraphics[width=\textwidth]{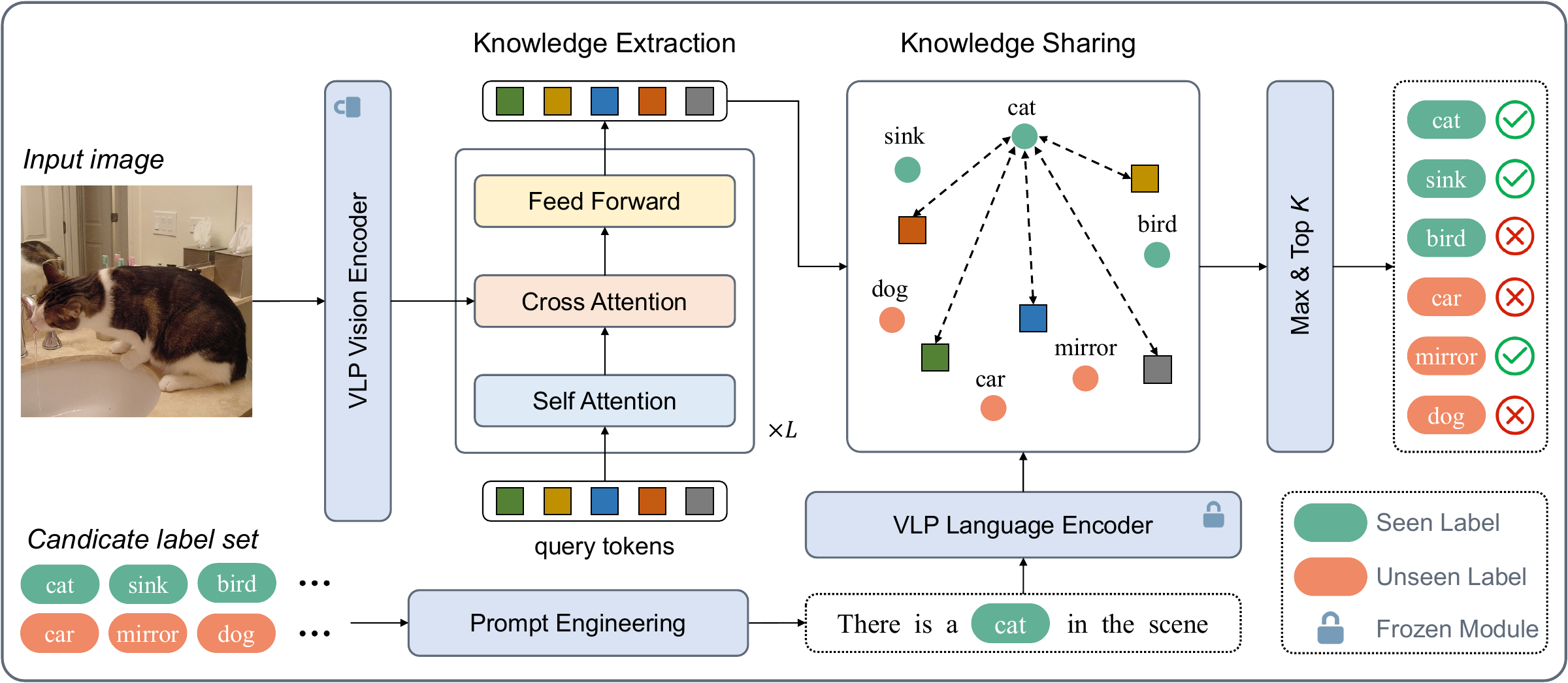}
	\caption{The detailed illustration of the proposed QKS framework. It takes a frozen VLP model as foundation followed by a knowledge extraction module and a knowledge sharing module. The former employs a set of label-agnostic query tokens to aggregate crucial and informative knowledge from the spatial features encoded by the VLP vision encoder, while the latter allows label embeddings encoded by the VLP language encoder to select tokens of interest as visual clues for recognition.}
	\label{fig:framework-detail}
\end{figure*}

As VLP models \cite{li2019visualbert,li2021align,radford2021learning,bao2022vlmo} evolve, open-vocabulary classification has served as an alternative for zero-shot prediction by transferring their image-text matching ability to the classification task. Latest open-vocabulary works in object detection \cite{du2022learning,gu2021open,zareian2021open,ma2022open} and image segmentation \cite{ghiasi2021open,huynh2022open}, together with knowledge distillation \cite{hinton2015distilling} and prompt tuning \cite{zhou2022learning}, have achieved impressive performance. In this context, research into the open-vocabulary label recognition has commenced. MKT \cite{he2023open} proposes a multi-modal knowledge transfer framework for exploring multi-modal knowledge in VLP models and a two-stream module for capturing both local and global features for multi-label task. However, redundant information in global features and incomplete objects in local features limit its performance. Also, beyond knowledge distillation and prompt tuning, a more effective pipeline for exploring multi-modal knowledge of VLP models remains to be established in terms of the open-vocabulary multi-label classification task.

\section{The Proposed Method}

\subsection{Problem Setting}
\label{sec:ps}
Providing that $\mathcal{X}$, $\mathcal{S}$ and $\mathcal{U}$ denote the image space, seen label set and unseen label set, respectively, where $\mathcal{S}$ and $\mathcal{U}$ are disjoint. Then, the training data $\mathcal{D}$ can be denoted as $\{(I_1,\mathcal{Y}_1),\cdots,(I_N,\mathcal{Y}_N)\}$, where $N$ denotes the number of training samples, $I_i\in\mathcal{X}$ is the image of the $i$-th training sample and $\mathcal{Y}_i\subseteq\mathcal{S}$ denotes the seen labels present in the image $I_i$. The goal of standard multi-label ZSL task is to learn a classifier based on the training set $\mathcal{D}$, such that the classifier can be well adapted to the identification of the unseen labels at test time, \textit{i.e.}, $f_\mathrm{ZSL}:\mathcal{X}\rightarrow\mathcal{U}$. Besides, a more challenging multi-label classification task of generalized zero-shot learning (GZSL) requires the learned classifier to recognize both the seen and unseen labels present in the test image, \textit{i.e.}, $f_\mathrm{GZSL}:\mathcal{X}\rightarrow\mathcal{S}\cup\mathcal{U}$.

\subsection{Overview}

Figure \ref{fig:framework-detail} illustrates the detailed pipeline of the proposed QKS framework. As shown, it mainly consists of a frozen VLP model and a knowledge extraction module as well as a knowledge sharing module. Concretely, the VLP model is frozen during the whole training phase to maintain its pretrained multi-modal knowledge, and its vision encoder and language encoder are employed to encode the spatial features and semantic embeddings for the input image and candidate labels, respectively. After that, the knowledge extraction module uses a fixed number of trainable label-agnostic query tokens to aggregate crucial and informative knowledge from the spatial features. The knowledge sharing module allows label embeddings to select tokens of interest as visual clues for recognition, enabling unseen label recognition in an open-vocabulary manner.

\subsection{Feature Extraction}
For brevity, we remove the subscript of input sample and denote it as $(I,\mathcal{Y})$. Then, the image $I$ is input into the VLP vision encoder to generate its spatial features, formulated as:
\begin{equation}
    \mathcal{F} = \Phi_\mathrm{v}^{\mathrm{VLP}}(I;\Theta_\mathrm{v}),
\end{equation}
where $\Phi_\mathrm{v}^{\mathrm{VLP}}$ denotes the visual encoder with $\Theta_\mathrm{v}$ being parameters; $\mathcal{F}\in\mathbb{R}^{H\times W\times C}$ and $H$, $W$ and $C$ are the height, width and the number of channels, respectively. After reshaping into a sequence of flatten 2D features, the spatial features are further mapped into a $d$-dimensional space, \textit{i.e.}, $\mathcal{F}\in\mathbb{R}^{HW\times d}$, by a linear projection layer.

\textbf{Prompt Pool for Label Embedding}. For label embedding, a common prompt technique is to populate a label into a handcrafted prompt template, such as ``There is a \{\} in the scene", which is subsequently fed into the VLP language encoder to obtain the label embedding. Due to the different training objectives between the VLP model and multi-label classification, prompt tuning \cite{he2023open} is introduced to improve label embeddings. However, experimental results show that it can only offer slight performance gain. 

In this work, we propose a prompt pool technique for label embedding, in which a set of prompt templates are carefully designed to serve as complete contexts for candidate labels. Specifically, we denote the prompt pool as $\{T_k\}_{k=0}^{K-1}$ with $K$ being the number of the pre-designed prompt templates, then the embedding of label $i$ is obtained as follows:
\begin{equation}
    \mathbf{t}_i=\frac{1}{K}\sum_{k=0}^{K-1}\Phi_\mathrm{t}^\mathrm{VLP}(\phi(T_k,i);\Theta_\mathrm{t}),
\end{equation}
where $\phi$ indicates the template filling operation; $\Phi_\mathrm{v}^{\mathrm{VLP}}$ denotes the VLP language encoder with $\Theta_\mathrm{t}$ being its parameters; $i\in\mathcal{S}\cup\mathcal{U}$ and  $\mathbf{t}_i\in\mathbb{R}^d$. Notably, compared to a single prompt template, our prompt pool is able to provide labels with richer and more diverse contexts, thus yielding robust label embeddings for matching with visual features.

\subsection{Knowledge Extraction}
We note that a label is semantically related to partial regions of an image. To extract semantically related regional features, a naive idea is to use label embeddings as queries to pool label-related features like Q2L \cite{liu2021query2label}. However, in the multi-label ZSL task, gathering features for unseen labels have unknown behaviors and could focus on irrelevant regions due to the lack of any training sample. Therefore, we propose to extract crucial vision knowledge by a fixed number of query tokens, which are trained to be label-agnostic and to focus on only relevant and informative regions. After fully gathering relevant visual features, these query tokens are subsequently shared across all labels and allow them to select parts of interest as visual clues for prediction.

Specifically, we design a set of label-agnostic query tokens, denoted as $\mathcal{Q}_0\in\mathbb{R}^{m\times d}$, where $m$ is the number of query tokens and $d$ denotes their dimensions. The query tokens are randomly initialized and trained to aggregate visual knowledge in the way of Transformer decoder. Taking the $l$-th layer as an example, the query tokens output by the previous layer are first input into a self-attention operation to exploit their self-correlation:
\begin{equation}
    \mathcal{Q}'_l = \mathcal{Q}_{l-1} + \mathrm{MSA}(\widetilde{\mathcal{Q}}_{l-1},\widetilde{\mathcal{Q}}_{l-1},\mathcal{Q}_{l-1}),
\end{equation}
where the tilde means the corresponding vectors modified by adding position encodings, and $\mathrm{MSA}(\cdot)$ denotes the multi-head self-attention, whose query, key and value matrices are all from the same source, \textit{i.e.}, query tokens. Then, the output tokens are input into the cross attention to gather informative knowledge from the spatial features:
\begin{equation}
    \mathcal{Q}''_l = \mathcal{Q}'_l + \mathrm{MSA}(\widetilde{\mathcal{Q}}'_l,\widetilde{\mathcal{F}},\mathcal{F}),
\end{equation}
where the query matrix of $\mathrm{MSA(\cdot)}$ is from query tokens, and the key and value matrices are both from the spatial features $\mathcal{F}$. After that, a feed-forward network $\mathrm{FFN}(\cdot)$ is employed to further update the query tokens:
\begin{equation}
    \mathcal{Q}_l = \mathcal{Q}''_l + \mathrm{FFN}(\mathcal{Q}''_l).
\end{equation}
Notably, as these processes proceed layer by layer, the final query tokens $\mathcal{Q}_L\in\mathbb{R}^{m\times d}$ fully gather crucial visual knowledge from the spatial features that is of great importance for label recognition. Here $L$ indicates the number of layers.


\subsection{Knowledge Sharing}
Once the critical and informative visual knowledge of spatial features is extracted, these query tokens are subsequently shared across all labels and allow them to select tokens of interest as important visual clues for recognition. Specifically, we first decouple the query tokens $\mathcal{Q}_L$ into a set of vectors $\{\mathbf{q}_j\}_{j=0}^{m-1}$ and $\mathbf{q}_j\in\mathbb{R}^d$. Then, the score for label $i$ appearing in the image $I$ is obtained by the maximum matching response of the label embedding $\mathbf{t}_i$ over the $m$ query vectors, which is formulated as follows:
\begin{equation}
    s_i = \max_{j=0,1,\cdots,n-1}\langle \mathbf{t}_i,\mathbf{q}_j \rangle,
\end{equation}
where $\langle\cdot,\cdot\rangle$ denotes inner product. In other words, the query token of most interest is regarded as the visual clue and evidence for the recognition of the label $i$. In this way, label recognition is decoupled into two independent phase: knowledge extraction for information filtering and knowledge sharing for semantic matching. The former focuses solely on extracting critical and informative regional features, while the latter is dedicated to selecting the most relevant features with respect to the labels for identification. Notably, such paradigm is particularly beneficial for the recognition of unseen labels.

\subsection{Ranking as Classification}
Recent works \cite{narayan2021discriminative,he2023open} commonly train models by ranking learning, which aims to rank the matching scores of image features with the positive label embeddings ahead, by a margin of at least 1, to that of the negative label embeddings. However, ranking learning generally works in a unit hyper-sphere, ignoring the magnitude of feature vectors, which is also important for label recognition since it reflects the strength information of features.

Having this in mind, we take the matching score $s_i$ without normalization operation and reformulate the ranking loss into a form of classification as follows:
\begin{equation}
    \mathcal{L} = -\sum_{p\in\mathcal{Y}}\log\sigma(s_p) - \sum_{n\notin\mathcal{Y}}\log(1-\sigma(s_n)),
\end{equation}
where the $\sigma(\cdot)$ is sigmoid function; $\mathcal{Y}$ is the set of seen label annotations for the image $I$, \textit{i.e.}, $\mathcal{Y}\subseteq\mathcal{S}$. Compared with the standard ranking loss, the classification loss $\mathcal{L}$ is actually a margin-agnostic ranking loss, which expects that the margin to be as large as possible, rather than just being satisfied with 1. On the other hand, unlike standard cross entropy that requires to learn classifiers, our loss $\mathcal{L}$ takes the label embeddings obtained by the VLP language encoder as classifiers. In this way, the angles between query tokens and label embeddings as well as their magnitudes are both taken into account for matching, thereby facilitating label recognition.

\section{Experiments}

\subsection{Experiment Setup}

\subsubsection{Datasets.} We evaluate our framework on two widely used benchmarks: NUS-WIDE \cite{chua2009nus} and Open Images \cite{kuznetsova2020open}. The \textbf{NUS-WIDE} dataset is a web dataset with 161,789 images for training and 107,859 for testing, covering 81 human verified labels as well as 925 labels acquired from Flickr user tags. As in \cite{huynh2020shared}, we treat the 925 labels as seen labels and remaining 81 labels as unseen labels. The \textbf{Open Images} (v4) is a large-scale dataset. It originally contains nearly 9 million training images as well as 41,620 and 125,456 images in validation and test sets. Similar to previous works \cite{huynh2020shared,he2023open}, we take 7,186 labels as seen labels, where each label has at least 100 images. As for unseen labels, the most frequent 400 test labels that are not in seen labels are selected. However, since some urls are dead, we were able to download only 7,987,856 and 110,066 images for training and testing, respectively. 


\subsubsection{Implementation Details.} Following the settings of MKT \cite{he2023open}, we choose the pre-trained CLIP \cite{radford2021learning} with  vision encoder being ViT-B/16 as our VLP model. For a fair comparison, the input images are resized into $224\times 224$ in both training and testing phases. Without any data augmentation, the whole framework is trained using AdamW optimizer \cite{loshchilov2017decoupled} with a batch size of 64. The learning rate is initialized as 1e-5 and decays by a factor of 10 when the loss plateaus. The number of query tokens $m$ and the number of layers $L$ of the vision knowledge extraction module are set as 12 and 7 for NUS-WIDE, as well as 22 and 8 for Open Images, respectively.


\subsection{Comparisons with State-of-the-arts}

\setlength{\aboverulesep}{2pt}
\setlength{\belowrulesep}{2pt}
\setlength{\abovetopsep}{0pt}
\renewcommand{\tabcolsep}{1pt}
\newcolumntype{C}{p{0.68cm}<{\centering}}

\begin{table}[t]
\small
\centering
\begin{tabularx}{\linewidth}{X<{\centering}p{0.8cm}<{\centering}CCCCCCC}
\toprule[0.5mm]
\multirow{2}{*}{\textbf{Method}} & \multirow{2}{*}{\textbf{Task}} & \multicolumn{3}{c}{\textbf{$\boldsymbol{K}$ = 3}} & \multicolumn{3}{c}{\textbf{$\boldsymbol{K}$ = 5}} & \multirow{2}{*}{\textbf{mAP}} \\ 
 & & \textbf{P} & \textbf{R} & \textbf{F1} & \textbf{P} & \textbf{R} & \textbf{F1} & \\
 \toprule[0.5mm]
\multirow{2}{*}{LESA \shortcite{huynh2020shared}} & ZSL & 25.7 & 41.1 & 31.6 & 19.7 & 52.5 & 28.7 & 19.4  \\
 & GZSL & 23.6 & 10.4 & 14.4 & 19.8 & 14.6 & 16.8 & 5.6 \\
 \cmidrule(lr{0.5em}){2-9}
 \multirow{2}{*}{ZS-SDL \shortcite{ben2021semantic}} & ZSL & 24.2 & 41.3 & 30.5 & 18.8 & 53.4 & 27.8 & 25.9 \\
  & GZSL & 27.7 & 13.9 & 18.5 & 23.0 & 19.3 & 21.0 & 12.1 \\
 \cmidrule(lr{0.5em}){2-9}
 \multirow{2}{*}{BiAM \shortcite{narayan2021discriminative}} & ZSL & 26.6 & 42.5 & 32.7 & 20.5 & 54.6 & 29.8 & 25.9 \\
  & GZSL & 25.2 & 11.1 & 15.4 & 21.6 & 15.9 & 18.2 & 9.4 \\
 \cmidrule(lr{0.5em}){2-9}
\multirow{2}{*}{CLIP \shortcite{radford2021learning}} & ZSL & 32.1 & 39.7 & 35.5 & 25.6 & 53.1 & 34.6 & 34.7 \\
 & GZSL & 33.2 & 14.6 & 20.3 & 27.4 & 20.2 & 23.2 & 16.8 \\
 \cmidrule(lr{0.5em}){2-9}
\multirow{2}{*}{DualCoOp \shortcite{sun2022dualcoop}} & ZSL & 37.3 & 46.2 & 41.3 & 28.7 & 59.3 & 38.7 & 43.6 \\
 & GZSL & 31.9 & 13.9 & 19.4 & 26.2 & 19.1 & 22.1 & 12.0 \\
 \cmidrule(lr{0.5em}){2-9}
\multirow{2}{*}{(ML)$^2$-Enc \shortcite{liu20232}} & ZSL & - & - & 32.8 & - & - & 32.3 & 29.4 \\
 & GZSL & - & - & 15.8 & - & - & 19.2 & 10.2 \\
 \cmidrule(lr{0.5em}){2-9}
\multirow{2}{*}{MKT \shortcite{he2023open}} & ZSL & 27.7 & 44.3 & 34.1 & 21.4 & 57.0 & 31.1 & 37.6 \\
 & GZSL & 35.9 & 15.8 & 22.0 & 29.9 & 22.0 & 25.4 & 18.3 \\
 \cmidrule(lr{0.5em}){2-9}
\multirow{2}{*}{\textbf{QKS}} & ZSL & \textbf{38.1} & \textbf{47.4} & \textbf{42.3} & \textbf{30.1} & \textbf{62.3} & \textbf{40.5} & \textbf{49.5} \\
 & GZSL & \textbf{39.9} & \textbf{17.4} & \textbf{24.2} & \textbf{33.6} & \textbf{24.4} & \textbf{28.3} & \textbf{22.5} \\
\bottomrule[0.5mm]
\end{tabularx}
\caption{Comparison of our QKS with existing methods on the NUS-WIDE dataset. All metrics, including precision (P), recall (R) and F1 score under the top-$K$ predictions and mean average precision (mAP) are in \% and the optimal scores are highlighted in bold.}
\label{tab:nuswide}
\end{table}

\begin{table}[t]
\small
\centering
\begin{tabularx}{\linewidth}{X<{\centering}p{0.8cm}<{\centering}CCCCCCC}
\toprule[0.5mm]
\multirow{2}{*}{\textbf{Method}} & \multirow{2}{*}{\textbf{Task}} & \multicolumn{3}{c}{\textbf{$\boldsymbol{K}$ = 10}} & \multicolumn{3}{c}{\textbf{$\boldsymbol{K}$ = 20}} & \multirow{2}{*}{\textbf{mAP}} \\ 
 & & \textbf{P} & \textbf{R} & \textbf{F1} & \textbf{P} & \textbf{R} & \textbf{F1} & \\
 \toprule[0.5mm]
\multirow{2}{*}{LESA \shortcite{huynh2020shared}} & ZSL & 0.7 & 25.6 & 1.4 & 0.5 & 37.4 & 1.0 & 41.7  \\
 & GZSL & 16.2 & 18.9 & 17.4 & 10.2 & 23.9 & 14.3 & 45.4 \\
 \cmidrule(lr{0.5em}){2-9}
 \multirow{2}{*}{ZS-SDL \shortcite{ben2021semantic}} & ZSL & 6.1 & 47.0 & 10.7 & 4.4 & 68.1 & 8.3 & 62.9 \\
  & GZSL & 35.3 & 40.8 & 37.8 & 23.6 & 54.5 & 32.9 & 75.3 \\
 \cmidrule(lr{0.5em}){2-9}
 \multirow{2}{*}{BiAM \shortcite{narayan2021discriminative}} & ZSL & 3.9 & 30.7 & 7.0 & 2.7 & 41.9 & 5.5 & 65.6 \\
  & GZSL & 13.8 & 15.9 & 14.8 & 9.7 & 22.3 & 14.8 & 81.7 \\
 \cmidrule(lr{0.5em}){2-9}
\multirow{2}{*}{CLIP \shortcite{radford2021learning}} & ZSL & 8.0 & 61.9 & 14.1 & 5.2 & 81.3 & 9.8 & 70.9 \\
 & GZSL & 17.8 & 20.6 & 19.1 & 12.7 & 29.5 & 17.7 & 77.3 \\
 \cmidrule(lr{0.5em}){2-9}
\multirow{2}{*}{(ML)$^2$-Enc \shortcite{liu20232}} & ZSL & - & - & 7.5 & - & - & 6.5 & 65.7 \\
 & GZSL & - & - & 27.6 & - & - & 24.1 & 79.9 \\
 \cmidrule(lr{0.5em}){2-9}
\multirow{2}{*}{MKT \shortcite{he2023open}} & ZSL & 11.1 & 86.8 & 19.7 & 6.1 & 94.7 & 11.4 & 68.1 \\
 & GZSL & 37.8 & 43.6 & 40.5 & 25.4 & 58.5 & 35.4 & 81.4 \\
 \cmidrule(lr{0.5em}){2-9}
\multirow{2}{*}{\textbf{QKS}} & ZSL & \textbf{11.8} & \textbf{91.9} & \textbf{20.9} & \textbf{6.2} & \textbf{97.2} & \textbf{11.7} & \textbf{72.6} \\
 & GZSL & \textbf{42.7} & \textbf{49.5} & \textbf{45.8} & \textbf{29.2} & \textbf{67.6} & \textbf{40.8} & \textbf{85.5} \\
\bottomrule[0.5mm]
\end{tabularx}
\caption{Comparison of our QKS with existing methods on the Open Images dataset. All metrics, including precision (P), recall (R) and F1 score under the top-$K$ predictions and mean average precision (mAP) are in \% and the optimal scores are highlighted in bold.}
\label{tab:openimage}
\end{table}

\subsubsection{Results on NUS-WIDE.} The left part of Table \ref{tab:nuswide} reports experimental results on NUS-WIDE dataset. As shown, our QKS framework achieves best performance on all metrics. Specifically, as an OV-based multi-label classification method, our QKS outperforms MKT by a remarkable margin in terms of mAP, reaching 11.9\% and 4.2\% on the ZSL and GZSL tasks, respectively. Besides, on another important F1 score, our QKS also shows pronounced merits on the ZSL task, improving the score from 34.1\% to 42.3\% for F1@3 and from 31.1\% to 40.5\% for F1@5, a total of 8.2\% and 9.4\% gains respectively. Similar merits also can be observed on the GZSL task. The all-around overwhelming advantages demonstrate the superiority of our open-vocabulary framework. In addition, the OV-based approaches generally perform much better than the ZS-based methods on both ZSL and GZSL tasks, highlighting the effectiveness of the multi-modal knowledge contained in the pre-trained VLP model for the recognition of seen and unseen labels. 



\subsubsection{Results on Open Images.} The right part of Table \ref{tab:openimage} shows experimental results on Open Images dataset. On the ZSL task, our QKS accomplishes best performance on all metrics. Especially in mAP, QKS raises mAP from 68.1\% to 72.6\% compared to MKT, a total of 4.5\% improvement. On the GZSL task, our QKS outperforms the MKT by 4.1\%, 5.4\% and 5.3\% in mAP, F1@10 and F1@20, respectively. Significant advantages on both tasks demonstrate the superiority of our open-vocabulary multi-label framework. It is worth noting that we actually have nearly 10\% fewer images than the official dataset due to some broken links. Our QKS are expected to perform better if the official Open Images are available for training.

\subsection{Ablation Study}

In this section, we conduct various ablation studies on the NUS-WIDE to evaluate the effectiveness of key designs.

\begin{table}[t]
\small
\centering
\begin{tabularx}{\linewidth}{X<{\centering}X<{\centering}p{1.2cm}<{\centering}p{1.0cm}<{\centering}p{1.8cm}<{\centering}p{1.8cm}<{\centering}}
\toprule[0.5mm]
$\mathbf{\Phi}_\mathbf{v}^\mathbf{VLP}$ & $\mathbf{\Phi}_\mathbf{t}^\mathbf{VLP}$ & \textbf{Task} & \textbf{mAP} & \textbf{F1 ($\boldsymbol{K}$ = 3)} & \textbf{F1 ($\boldsymbol{K}$ = 5)} \\
 \toprule[0.5mm]
\multirow{2}{*}{\Checkmark} & \multirow{2}{*}{\Checkmark} & ZSL & 33.7 & 34.9 & 33.2 \\
 & & GZSL & 19.6 & 21.4 & 25.1 \\
 \cmidrule(lr{0.5em}){3-6}
\multirow{2}{*}{\XSolidBrush} & \multirow{2}{*}{\Checkmark} & ZSL & 38.4 & 35.6 & 34.4 \\
 & & GZSL & 19.1 & 20.6 & 24.3 \\
 \cmidrule(lr{0.5em}){3-6}
\multirow{2}{*}{\Checkmark} & \multirow{2}{*}{\XSolidBrush} & ZSL & 43.3 & 38.0 & 36.9 \\
 & & GZSL & 20.1 & 22.1 & 25.8 \\
 \cmidrule(lr{0.5em}){3-6}
\multirow{2}{*}{\XSolidBrush} & \multirow{2}{*}{\XSolidBrush} & ZSL & \textbf{49.5} & \textbf{42.3} & \textbf{40.5} \\
 & & GZSL & \textbf{22.5} & \textbf{24.2} & \textbf{28.3} \\
\bottomrule[0.5mm]
\end{tabularx}
\caption{Effect of the VLP model on the performance of our QKS. The check mark means to unfreeze the corresponding encoder, while the cross mark indicates the opposite.}
\label{tab:vlp}
\end{table}

\subsubsection{Effect of the VLP model.} In order to figure out the effect of the VLP model, we design several variants of the proposed QKS, which either individually or simultaneously unfreeze the VLP vision encoder and the VLP language encoder during training. Experimental results are reported in Table \ref{tab:vlp}. When the language encoder is unfrozen, the variants perform poorly in recognizing unseen labels, regardless of whether the visual encoder is also unfrozen or not. We argue this is due to the inability to transfer knowledge between seen and unseen labels due to parameter tuning of the language encoder. When only the visual encoder is unfrozen, the whole framework degenerates into a traditional zero-shot model that only exploits single-modal knowledge from the language encoder. By freezing both encoders, the multi-modal knowledge of the VLP model learned from large-scale image-text pairs is preserved, and the QKS model achieves significant performance boost compared to its variants on both tasks, evincing the effectiveness of our open-vocabulary framework.

\begin{table}[t]
\small
\centering
\begin{tabularx}{\linewidth}{X<{\centering}p{1.2cm}<{\centering}p{1.5cm}<{\centering}p{1.8cm}<{\centering}p{1.8cm}<{\centering}}
\toprule[0.5mm]
\textbf{Prompt} & \textbf{Task} & \textbf{mAP} & \textbf{F1 ($\boldsymbol{K}$ = 3)} & \textbf{F1 ($\boldsymbol{K}$ = 5)} \\
 \toprule[0.5mm]
\multirow{2}{*}{Single} & ZSL & 47.4 & 40.5 & 38.0 \\
 & GZSL & 20.9 & 23.2 & 27.1 \\
 \cmidrule(lr{0.5em}){2-5}
\multirow{2}{*}{Tuning} & ZSL & 47.8 & 41.6 & 39.2 \\
 & GZSL & 21.1 & 23.4 & 27.4 \\
 \cmidrule(lr{0.5em}){2-5}
\multirow{2}{*}{Pool} & ZSL & \textbf{49.5} & \textbf{42.3} & \textbf{40.5} \\
 & GZSL & \textbf{22.5} & \textbf{24.2} & \textbf{28.3} \\
\bottomrule[0.5mm]
\end{tabularx}
\caption{Effect of single prompt, prompt tuning and prompt pool on the performance of the proposed QKS.}
\label{tab:prompt}
\end{table}

\subsubsection{Effect of the prompt pool.}
To explore the effectiveness of our prompt pool, we also conduct experiments on our QKS with single prompt and prompt tuning \cite{he2023open}. As shown in Table \ref{tab:prompt}, our prompt pool achieves best performance across all metrics on both ZSL and GZSL tasks. Specifically, as far as the ZSL task is concerned, compared with single prompt, prompt tuning only improves mAP, F1@3 and F1@5 by 0.4\%, 1.1\% and 1.2\%, while our prompt pool raises these metrics by 2.1\%, 1.8\% and 2.5\%, respectively. Similar phenomena can also be observed on the GZSL task. Notably, our prompt pool performs better in terms of the prompt engineering for the open-vocabulary multi-label classification task.

\begin{table}[t]
\small
\centering
\begin{tabularx}{\linewidth}{X<{\centering}p{1.2cm}<{\centering}p{1.0cm}<{\centering}p{1.8cm}<{\centering}p{1.8cm}<{\centering}}
\toprule[0.5mm]
\textbf{Objective} & \textbf{Task} & \textbf{mAP} & \textbf{F1 ($\boldsymbol{K}$ = 3)} & \textbf{F1 ($\boldsymbol{K}$ = 5)} \\
 \toprule[0.5mm]
\multirow{2}{*}{Ranking} & ZSL & 49.0 & 37.3 & 36.0 \\
 & GZSL & 21.9 & 23.8 & 27.6 \\
 \cmidrule(lr{0.5em}){2-5}
\multirow{2}{*}{Classification} & ZSL & \textbf{49.5} & \textbf{42.3} & \textbf{40.5} \\
 & GZSL & \textbf{22.5} & \textbf{24.2} & \textbf{28.3} \\
\bottomrule[0.5mm]
\end{tabularx}
\caption{Effect of ranking learning and classification learning on the performance of the proposed QKS framework.}
\label{tab:loss}
\end{table}

\subsubsection{Effect of the objective function.} We also compare the performance of classification learning and ranking learning. As shown in Table \ref{tab:loss}, classification learning achieves better scores than ranking learning over all metrics on both tasks. Particularly, in terms of the ZSL task, classification learning has a remarkable advantage in F1 scores, surpassing ranking learning by 5.0\% and 3.5\% on F1@3 and F1@5, respectively. Notably, classification learning is more conducive to improving the precision and recall of the model on unseen label prediction. Besides, it is worth noting that with the same ranking learning, our QKS outperforms MKT by a considerable margin across all metrics, further confirming the superiority of our QKS framework.



\begin{figure}[t]
    \centering
    \subfigure[Variation of $m$]{
        \includegraphics[width=0.47\linewidth]{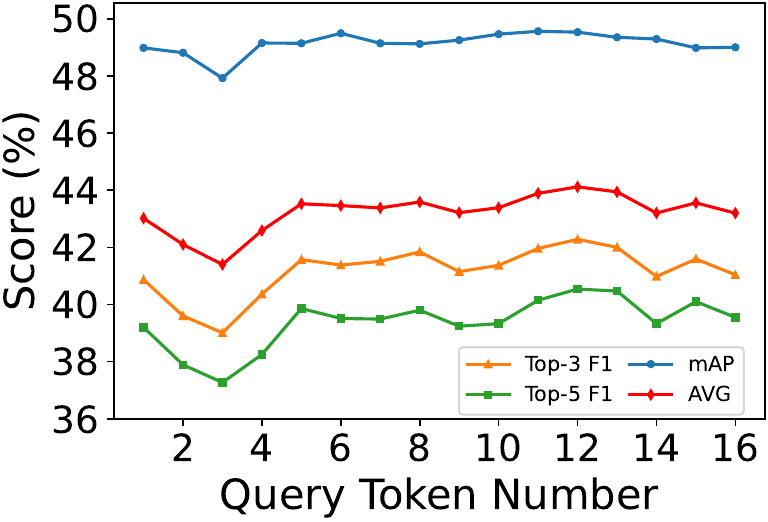}
        \label{fig:query}
    }
    \subfigure[Variation of $L$]{
        \includegraphics[width=0.47\linewidth]{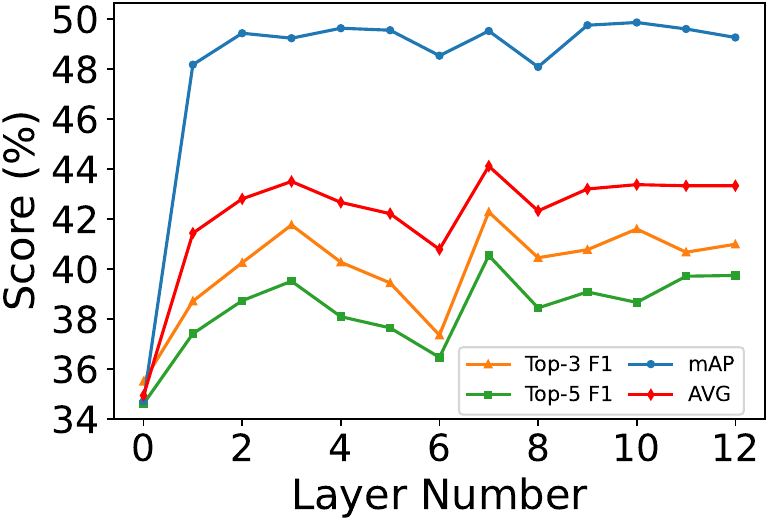}
        \label{fig:layer}
    }
    \caption{Effect of the hyper-parameters. The AVG (red curve) is the average score of mAP, Top-3 F1 and Top-5 F1.}
\end{figure}

\begin{figure*}[t]
	\centering
	\includegraphics[width=\textwidth]{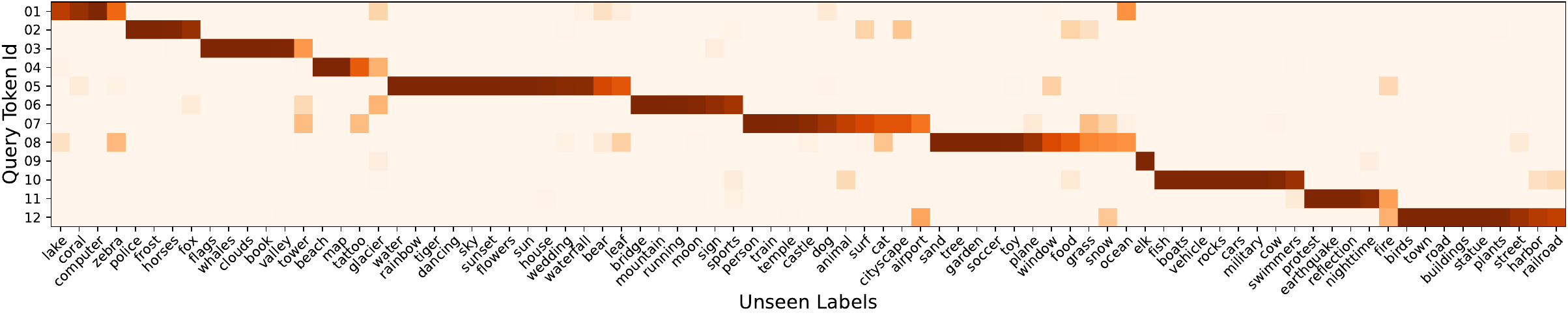}
	\caption{Visualization of the distribution of 81 unseen labels' preferences for 12 query tokens on the NUS-WIDE testing set.}
	\label{fig:heatmap}
\end{figure*}

\begin{figure}[t]
    \centering
    \subfigure[Distribution on GZSL task]{
        \includegraphics[width=0.47\linewidth]{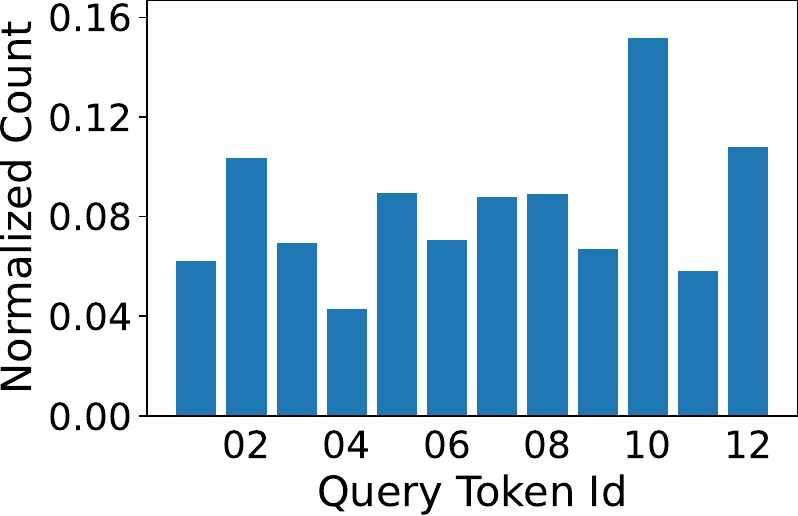}
    }
    \subfigure[Distribution on ZSL task]{
	  \includegraphics[width=0.47\linewidth]{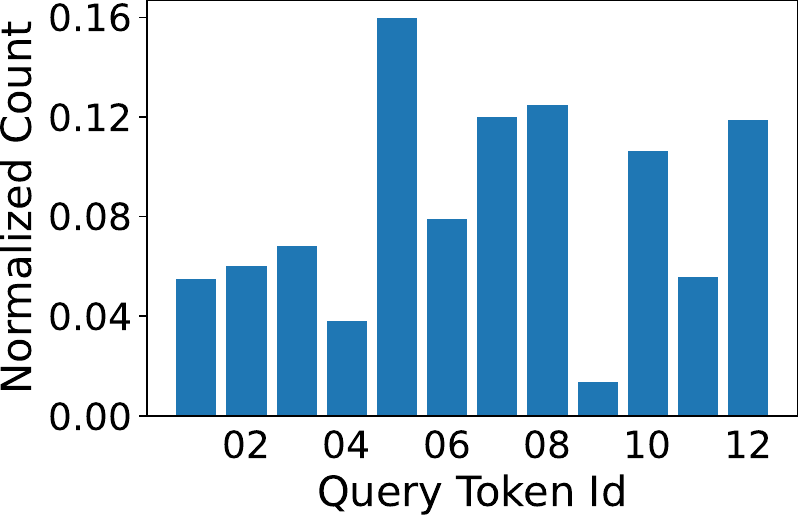}
    }
    \caption{Distribution of the number of positive labels with the maximum matching score for each query token.}
    \label{fig:distribution}
\end{figure}

\subsubsection{Effect of the knowledge extraction.} Here we investigate the effect of the knowledge extraction module. First we focus on how the metrics including mAP, F1@3 and F1@5 vary with the number of query tokens. As shown in Figure \ref{fig:query}, all curves exhibit a downward trend when $m\leq 2$, and then gradually climb to a stable state with some slight fluctuations as $m$ increases. Obviously, the QKS framework suffers at the initial stage as there are too few query tokens to fully extract visual knowledge for matching with label embeddings. In subsequent stages, all metrics are at a high level and reach their average maximum at $m=12$. 

On the other hand, the effect of the number of layers in the knowledge extraction module is shown in Figure \ref{fig:layer}. Note that the knowledge extraction module is removed and the QKS framework degenerates to the original CLIP when $L$ is $0$. Notably, our knowledge extraction module implements considerable gains on all metrics, even with $L$ being $1$. And when $L$ is $7$, our QKS reaches the maximum average score. However, the performance of QKS decreases as the number of layers continues to increase. We attribute this phenomenon to the limited data size, which makes the QKS prone to over-fitting as the number of parameters increases. Overall, the knowledge extraction module shows a powerful ability in acquiring critical and diverse knowledge from images for label recognition without relying on extra supervision signals.

\subsubsection{Effect of the knowledge sharing.} To check whether the knowledge sharing module works as expected, we first visualize the preference distribution of 81 unseen labels to query tokens on NUS-WIDE testing set. As shown in Figure \ref{fig:heatmap}, it is clear that most labels only favor a particular one of the 12 query tokens and each query token is shared by a group of unseen labels, which caters to our motivation and justifies our QKS framework. On the other hand, Figure \ref{fig:distribution} counts the frequency distribution of positive labels with the maximum matching score for each query token. On the GZSL task, we can observe a generally uniform distribution of frequencies with some fluctuations. Although the fluctuations intensify slightly on the ZSL task, most query tokens still make a high-frequency contribution to unseen label recognition. Evidently, our knowledge sharing module demonstrates to be effective for labels to select query tokens of interest, achieving significant performance improvement in unseen label recognition.



\begin{figure}[t]
	\centering
	\includegraphics[width=\linewidth]{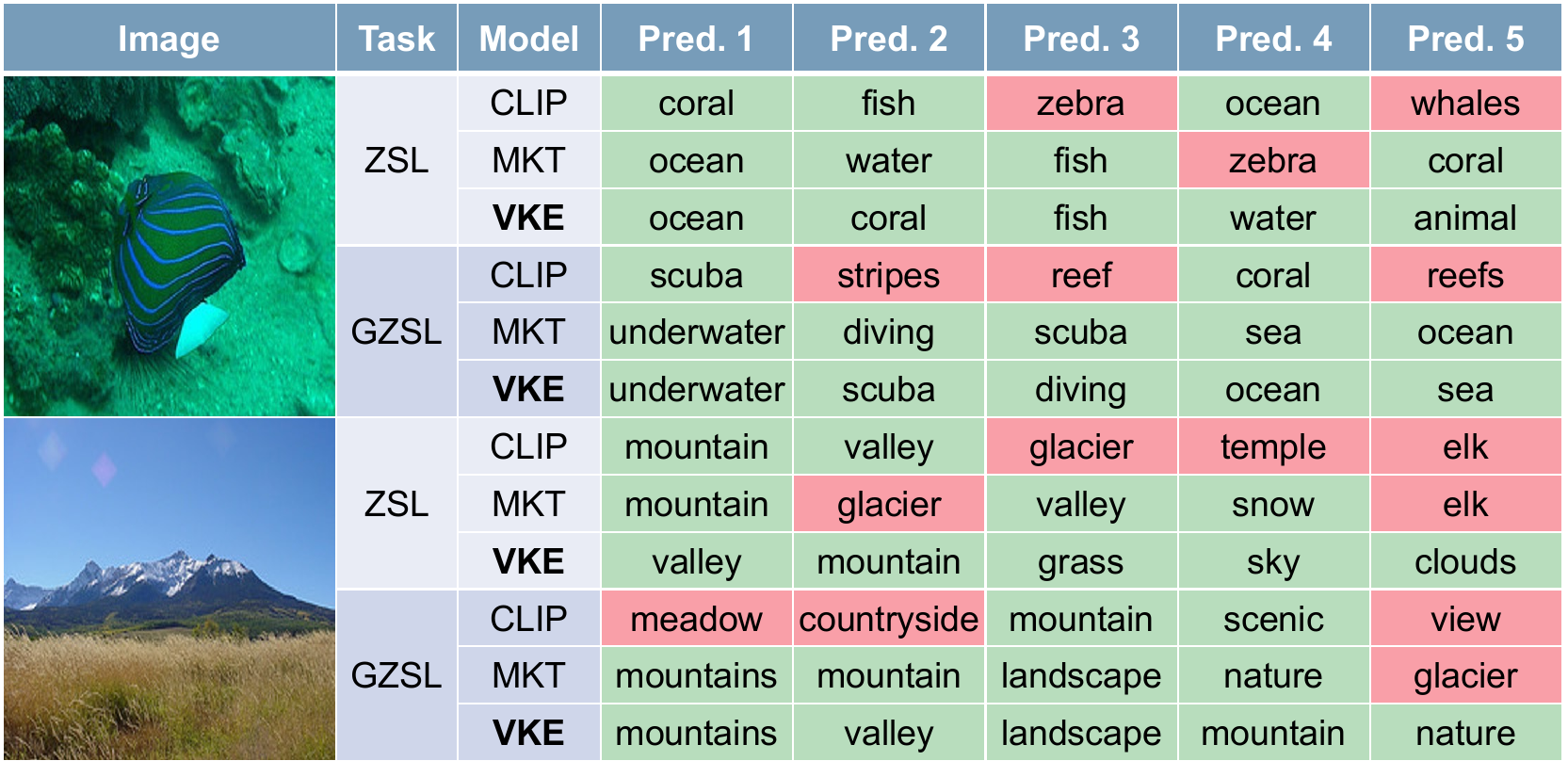}
	\caption{Comparison on the top-$5$ predictions of several open-vocabulary methods. Correct and incorrect predictions are distinguished by green and red cells, respectively. }
	\label{fig:prediction}
\end{figure}

\begin{figure}[t]
	\centering
	\includegraphics[width=\linewidth]{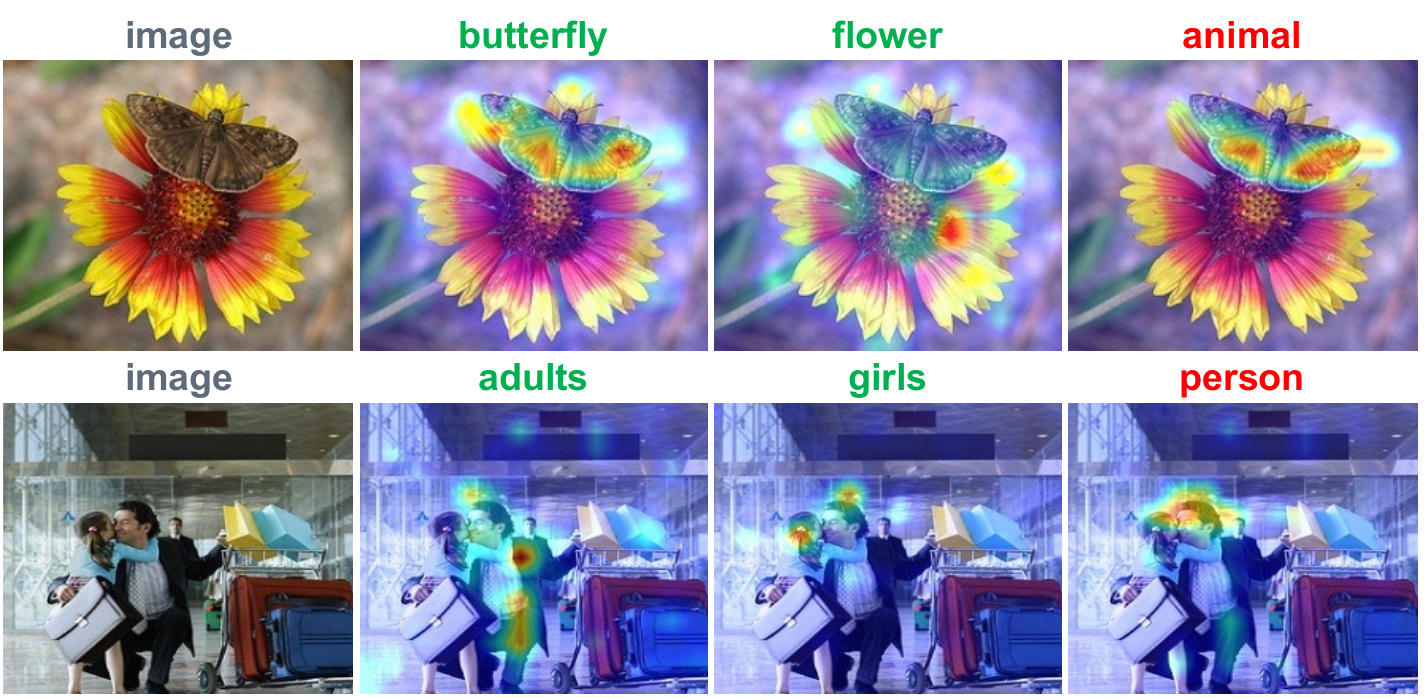}
	\caption{Visualization of attention maps for several images from the NUS-WIDE testing set. Seen and unseen labels are highlighted in green and red, respectively.}
	\label{fig:attention}
\end{figure}

\subsection{Case Study}
In this section, we qualitatively analyze the QKS framework by visualizing its prediction and attention maps on several images. Figure \ref{fig:prediction} compares the top-$5$ predictions of CLIP and MKT as well as our QKS on the ZSL and GZSL tasks. Overall, our QKS provides more precise and diverse predictions on both tasks compared to the other two OV-based methods, which more or less make wrong predictions. We also visualize the attention map of labels on images, which is obtained by averaging the attention weights over multiple heads for the query token with the maximum matching score. As shown in Figure \ref{fig:attention}, our QKS pinpoints semantically relevant regions for both seen and unseen labels. Specifically, the QKS shows an impressive ability to distinguish between labels with similar semantics. For example, the QKS focuses on face-related regions for \textit{person}, while highlighting regions such as clothing and braids for \textit{adults} and \textit{girls}, respectively. Overall, the proposed QKS is capable of capturing crucial and informative visual clues for both seen and unseen label recognition.

\section{Conclusion}
In this paper, we propose an advanced vision knowledge extraction framework for open-vocabulary multi-label classification task. Concretely, it takes the frozen CLIP model as the foundation of the whole framework to leverage its pretrained multi-modal knowledge, then employs a set of label-agnostic query tokens to extract crucial and informative visual clues, and a knowledge sharing module to allow candidate labels . We also modify ranking learning as a form of classification to further boost the performance of our proposed framework. Extensive experiments demonstrate the superiority of the proposed framework, and reasonable ablation studies prove the effectiveness of our key designs.

\bibliographystyle{named}
\bibliography{ijcai24}

\end{document}